\documentclass{article}
\usepackage{ijcai25}

\usepackage{times}
\usepackage{soul}
\usepackage{url}
\usepackage[hidelinks]{hyperref}
\usepackage[utf8]{inputenc}
\usepackage[small]{caption}
\usepackage{graphicx}
\usepackage{amsmath}
\usepackage{amsthm}
\usepackage{booktabs}
\usepackage{algorithm}
\usepackage{algorithmic}
\usepackage[switch]{lineno}

\newcommand{\norm}[1]{\left \lVert #1 \right \rVert}

\title{Framework GNN-AID: Graph Neural Network Analysis Interpretation and Defense}

\author{
Kirill Lukyanov$^{1,2,3}$
\and
Mikhail Drobyshevskiy$^{1,2}$\and
Georgii Sazonov$^{2,3}$\and \\
Mikhail Soloviov$^{2}$\and
Ilya Makarov$^{1,4}$\\
\affiliations
$^1$ ISP RAS Research Center for Trusted Artificial Intelligence, 109004 Moscow, Russia\\
$^2$ Ivannikov Institute for System Programming of the Russian Academy of Sciences, 109004 Moscow, Russia\\
$^3$ Moscow Institute of Physics and Technology (National Research University), 141700 Moscow, Russia\\
$^4$ Lomonosov Moscow State University, Leninskie Gory, 1, Moscow, 119991, Russia\\
$^5$ AIRI, 121170 Moscow, Russia\\
\emails
\{lukyanov.k, drobyshevsky, sazonovg\}@ispras.ru,
m.solovev@phystech.edu, iamakarov@hse.ru 
}

\begin{document}

\maketitle

\begin{abstract}
 The growing need for Trusted AI (TAI) highlights the importance of interpretability and robustness in machine learning models. However, many existing tools overlook graph data and rarely combine these two aspects into a single solution. Graph Neural Networks (GNNs) have become a popular approach, achieving top results across various tasks. We introduce GNN-AID (Graph Neural Network Analysis, Interpretation, and Defense), an open-source framework designed for graph data to address this gap. Built as a Python library, GNN-AID supports advanced trust methods and architectural layers, allowing users to analyze graph datasets and GNN behavior using attacks, defenses, and interpretability methods.

GNN-AID is built on PyTorch-Geometric, offering preloaded datasets, models, and support for any GNNs through customizable interfaces. It also includes a web interface with tools for graph visualization and no-code features like an interactive model builder, simplifying the exploration and analysis of GNNs. The framework also supports MLOps techniques, ensuring reproducibility and result versioning to track and revisit analyses efficiently.

GNN-AID is a flexible tool for developers and researchers. It helps developers create, analyze, and customize graph models, while also providing access to prebuilt datasets and models for quick experimentation. Researchers can use the framework to explore advanced topics on the relationship between interpretability and robustness, test defense strategies, and combine methods to protect against different types of attacks. 

We also show how defenses against evasion and poisoning attacks can conflict when applied to graph data, highlighting the complex connections between defense strategies.

GNN-AID is available at \href{https://github.com/ispras/GNN-AID}{github.com/ispras/GNN-AID}
 
\end{abstract}

\section{Introduction}
\label{introduction}

In recent years, trustworthiness has emerged as a critical area of research in artificial intelligence. It encompasses key aspects such as interpretability and security. Numerous methods for interpreting machine learning models have been proposed for various data domains, including images, text, and tabular data~\cite{burkart2021survey}. At the same time, new attack strategies targeting machine learning models, as well as corresponding defense mechanisms, are continually being developed~\cite{tian2022comprehensive}.

Access to diverse tools within a unified interface is highly advantageous for a machine learning developer. This allows for the application of multiple techniques without the need to search for and integrate separate implementations. To address this, several frameworks and toolboxes have been introduced that combine model interpretability, attack methods, defense mechanisms, and other functionalities. However, these frameworks often pay little attention to the graph data domain despite the rapid growth and active development of GNNs in recent years.

To fill this gap, we present GNN-AID, a tool fully dedicated to Graph Neural Networks. GNN-AID provides a comprehensive set of tools for \underline{a}nalyzing, \underline{i}nterpreting, attacking, and \underline{d}efending GNN models. It includes an extensible Python library complemented by a frontend module that offers most of the functionality through an intuitive interface. In this paper, we provide the following. Section \ref{sec:background} gives a background and reviews alternative tools, in section \ref{sec:system_overview} we describe the system architecture and its key components. Section \ref{sec:methods} outlines the implemented tools for interpretation, attacks, and defenses, and section \ref{sec:use_cases} presents the main usage scenarios for the framework, followed by a conclusion section \ref{sec:conclusion} discussing future directions.

\section{Background}
\label{sec:background}

In this section, we give a background on GNN models and methods for interpretation, attack, and defense and overview existing tools and libraries.

\subsection{Graph neural networks and trusted AI methods}

Graph Neural Networks started to gain significant attention in 2016, when the first convolutional layers --- GCN~\cite{gcn}, SAGE~\cite{graphsage}, GAT~\cite{gat}, and GIN~\cite{xu2018powerful} later --- were introduced. These architectures demonstrated state-of-the-art (SOTA) performance across various tasks, making them the most popular choices for graph-based problems. Typical tasks in the graph domain include node classification, link prediction, graph classification, graph generation, graph similarity measurement, and others. Usually, GNN architectures consist of convolutional layers that learn vector representations of a graph and/or its nodes, followed by a classifier that solves the target task. The specificity of GNNs from the models working with other data types lies in these convolutional layers.

Interpretation methods for machine learning models can be categorized into \textit{post-hoc} methods, which provide explanations after model training, and \textit{self-interpretable} methods, where explanations are built-in and generated during the training and included as part of the model’s output. For example, in node classification tasks, a local post-hoc explanation method would identify important graph elements (e.g., specific nodes or edges) within the neighborhood of a given node that mostly influenced the model’s prediction. Self-interpretable models often provide a broader range of explanation types. For instance, NeuronAnalysis~\cite{Han2022GlobalCI} in global interpretation can associate logical expressions with neurons in the final layer to describe their role in the model. However, there are very few such methods in the literature.
Another approach is counterfactual interpretation, which involves an expert interacting with a model by posing "what-if" questions and observing changes in its behavior. While these techniques are promising, GNN-AID does not yet support them.

Attacks on machine learning models can be broadly categorized into several types: \textit{evasion attacks} \cite{goodfellow2014explaining}, which aim to mislead models by manipulating inputs at inference time; \textit{poisoning attacks} \cite{zhang2022unsupervised}, where adversaries compromise the training process by introducing malicious data, including \textit{backdoor attacks} \cite{zheng2023motif}, where hidden triggers are implanted to induce specific behaviors during inference; \textit{privacy attacks} \cite{shaikhelislamov2023study}, which include attempts to infer sensitive information about the training data, such as Membership Inference (MI) attacks; \textit{model stealing attacks} \cite{oliynyk2023know}, where adversaries attempt to replicate or extract a model’s architecture and parameters for unauthorized use; and \textit{interpretation attacks} \cite{wang2022adversarial}, which aim to manipulate or mislead interpretability mechanisms, reducing trust in the model’s explanations.

Now, the framework supports post-hoc interpretation and self-interpretable methods, both evasion and poisoning attacks, along with corresponding defense methods to enhance the robustness of GNNs. Other types of attacks, such as privacy attacks, model stealing, and interpretation attacks, are not yet supported, but architectural extensions are already in place to facilitate the addition of privacy-related attacks, such as Membership Inference (MI) attacks, which are planned for implementation shortly.

\subsection{Trusted AI frameworks}
From the perspective of open source tools and frameworks, existing solutions can be broadly divided into those aimed at ensuring the trustworthiness of AI models across various input data types, typically images, text, and tabular data, and those specifically focused on graph data processing.

\subsubsection{Frameworks for trustworthiness}
Several frameworks are designed to ensure the trustworthiness of AI models across various domains. There are ones focused on interpretability, another incorporates attack and defense methods.
Most popular tools for explainable AI are designed for images, text and tabular data and do not support graph neural networks: AI Explainability 360 (Python)~\cite{arya2021ai}, Google Cloud’s AI Explanations~\cite{}, Python ELI5\footnote{\url{https://github.com/eli5-org/eli5}}, InterpretML~\cite{nori2019interpretml}, IML~\cite{molnar2018iml}, DeepExplain~\cite{ancona2017towards}, omnixai~\cite{wenzhuo2022-omnixai}.
Limited support of GNNs is present in Microsoft AzureML-Interpret\footnote{\url{https://docs.microsoft.com/python/api/overview/azure/ml/?view=azure-ml-py}}.

Frameworks focusing on attacks and defenses include CleverHans~\cite{papernot2018cleverhans}, Foolbox~\cite{rauber2017foolbox}, and ART~\cite{nicolae2018adversarial}. They can support GNNs but their methods need adaptation for graph data.

\subsubsection{Graph-specific libraries}
Now we consider libraries specifically developed for GNN analysis that include some TAI functionality. Table~\ref{tab:gnn-tools} gives a comparison.
Probably, the most popular one is PyTorch Geometric (PyG)~\cite{fey2019fast}: a widely-used library for building GNNs, offering datasets and prebuilt models. It provides several post-hoc local interpretation algorithms.
Deep Graph Library (DGL)~\cite{wang2019deep}: A flexible and scalable GNN framework that supports multiple backends including PyTorch. It implements only one interpretation method. 
DIG~\cite{JMLR:v22:21-0343}: A comprehensive library focusing on various GNNs tasks and includes an explainability tools package with a more diverse set of algorithms.
DeepRobust~\cite{li2020deeprobust} and GreatX~\cite{li2022recent} are focused on attacks and defenses on graph data and provide a bunch of methods.
GraphFramEx~\cite{amara2022graphframex} and CLARUS~\cite{metsch2024clarus} are designed specifically for GNN interpretability. While CLARUS offers certain interpretation methods and has a web page with interactive design, its codebase has not been maintained since 2022.

\begin{table}[h]
\centering
\begin{tabular}{p{1.6cm}cccc}
\toprule
\textbf{Name} & \textbf{Interpretation} & \textbf{Attacks} & \textbf{Defenses} & \textbf{GUI} \\ \midrule
PyG & 5  & - & -  & -  \\
DGL & 1  & - & -  & -  \\
DIG & 8  & - & -  & -  \\
DeepRobust& -  & 12 & 13 & -  \\
GreatX & - & 18 & 17 & - \\
GraphFramEx & 15 & - & - & - \\
CLARUS & 4  & - & -  & + \\
\textbf{GNN-AID} & 8  & 7 & 7 & + \\
\bottomrule
\end{tabular}
\caption{Comparison of tools for GNNs analysis with numbers of corresponding algorithms.}
\label{tab:gnn-tools}
\end{table}

While there are many tools available, none provides a comprehensive set of functionalities for interpretation, attacks, and defenses tailored specifically for graph data. Moreover, very few tools include a user interface, which significantly lowers the entry barrier for beginners. GNN-AID aims to address these gaps by offering an integrated, extensible framework with a focus on usability and functionality for graph-based machine learning.

\section{System Overview}
\label{sec:system_overview}

\subsection{Architecture and pipelines}

\begin{figure}[h]
    \centering
    \includegraphics[width=0.48\textwidth]{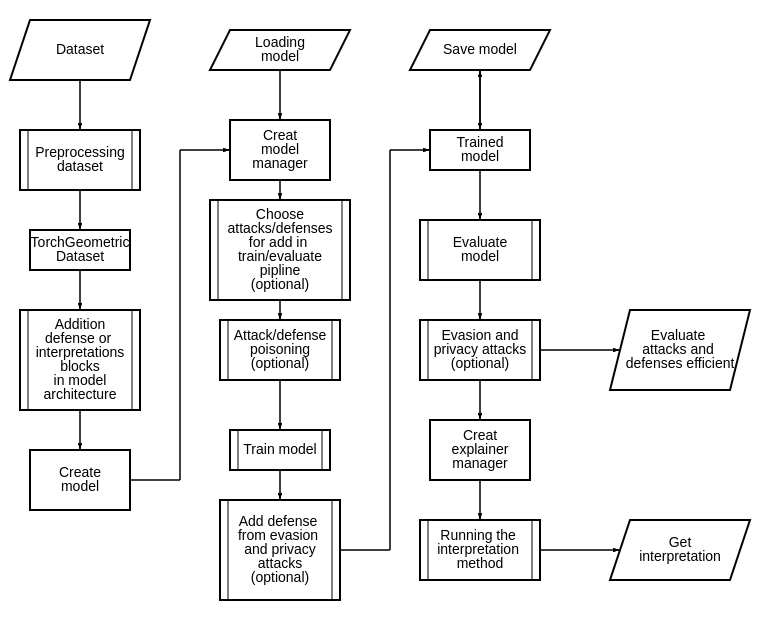}
    \caption{GNN-AID main pipleine. }
\label{fig:GNN-AID-pipeline}
\end{figure}

The main general framework pipeline can be seen in the diagram Figure \ref{fig:GNN-AID-pipeline} Additional information about these steps is provided below.

\subsubsection{Dataset processing module}

The first step in using GNN-AID is dataset selection. It consists of two stages: selecting a graph or a set of graphs (for graph classification tasks) and defining the dataset's variable components: features, labels, and the target task.

This separation accounts for the flexibility of solving multiple tasks on the same graph. For instance, a graph can support different feature construction methods, labeling schemes, or task definitions, such as switching from node classification to edge prediction. Attributes (e.g., a person’s age in a social graph or atom types in a molecular graph) are distinguished from features, which are numerical representations derived from these attributes. 
\subsubsection{Model, model manager, attacks, and defenses}

The model is constructed as a modular system using universal building blocks that may include the following components:  
\begin{itemize}
    \item A graph convolutional layer or a fully connected layer
    \item A batch normalization layer
    \item An activation function
    \item A dropout layer
\end{itemize}

For node classification tasks, pooling layers must be placed between blocks to define the representations of nodes and the entire graph. In addition to sequential connections, blocks can also be linked using skip connections.

The model manager, in turn, is responsible for training the model, saving it, loading it, performing attacks on the model, and applying defense methods. The process of training, conducting attacks, and applying defenses consists of the following stages illustrated in Algorithm \ref{alg:training_pipeline} 

\begin{algorithm}[tb]
    \caption{Train and Evaluate Pipeline with Defenses and Attacks}
    \label{alg:training_pipeline}
    \begin{algorithmic}[1]
        \REQUIRE Initial dataset $\mathcal{D}$, its training part $\mathcal{D}_{train}$, dataset's part chosen by boolean mask $\mathcal{D}_{mask}$, model $\mathcal{M}$ with parameters $\Theta$, number of epochs $N$, loss after train $L$, batch size $B$, early stopping criteria, optimization algorithm, all attacks and defenses methods are used if they were chosen 
        \STATE $\mathcal{D}_{train} \gets \text{PoisonAttack}(\mathcal{D}_{train})$
        \COMMENT{Applying poison attacks to the training part of the dataset}
        \STATE $\mathcal{D}_{train} \gets \text{PoisonDefense}(\mathcal{D}_{train})$
        \COMMENT{Applying poison defense to mitigate poison attack}
        \STATE Initialize $\mathcal{M}$ with parameters $\Theta_0$
        \COMMENT{Model initialization}
        \FOR{$epoch = 1$ to $N$}
            \STATE $\mathcal{B} \gets \text{SplitIntoBatches}(\mathcal\mathcal{D}_{train}, B)$
            \FOR{each $b \in \mathcal{B}$}
                \STATE $\mathcal{M}, b, b* \gets \text{PrivacyDefense}(\mathcal{M}, b)$
                \COMMENT{Add defense against privacy attacks before training on batch}
                \STATE $\mathcal{M}, b, b* \gets \text{EvasionDefense}(\mathcal{M}, b, b*)$
                \COMMENT{Add defense against evasion attacks before train on batch}
                \STATE $\mathcal{M}, \mathcal{L} \gets \text{Train on batch}(\mathcal{M}, b)$
                \COMMENT{Train model on the current batch}
                \STATE $\mathcal{M}, \mathcal{L} \gets \text{PrivacyDefense}(\mathcal{M}, \mathcal{L})$
                \COMMENT{Add defense against privacy attacks after training on batch}
                \STATE $\mathcal{M}, \mathcal{L} \gets \text{EvasionDefense}(\mathcal{M}, \mathcal{L})$
                \COMMENT{Add defense against evasion attacks after training on batch}
                \STATE $\Theta \gets \text{OptimizationStep}(\mathcal{M}, \mathcal{L}, \Theta)$
                \IF{$\text{EarlyStoppingCriteriaMet}()$}
                    \STATE \textbf{break}
                \ENDIF
            \ENDFOR
        \ENDFOR
        \STATE $\mathcal{D}_{mask} \gets \text{EvasionAttacks}(\mathcal{M}, \mathcal{D}_{mask})$
        \COMMENT{Applying evasion attack}
        \STATE $\mathcal{M} \gets \text{PrivacyAttacks}(\mathcal{M}, \mathcal{D}_{mask})$
        \COMMENT{Applying privacy attack}
        \STATE $Metrics \gets \text{EvaluateModel}(\mathcal{M}, \mathcal{D}_{mask})$ 
        \COMMENT{Evaluate model on $\mathcal{D}_{mask}$}
        \STATE \textbf{return} $\mathcal{M}$, $Metrics$
    \end{algorithmic}
\end{algorithm}

It should be noted that, in addition to the listed actions, user-defined hooks are supported to modify the model's behavior and/or the training process. 

Defenses against privacy attacks and evasion attacks are executed before and/or after training on a batch since most of these methods require data modification (for example, Adversarial Training) or gradient modification (for example,  Gradient Regularization). Additionally, some methods modify the model's architecture itself; such modifications are performed once before training begins unless otherwise specified by the defense method. 

Each attack and defense method can be defined or omitted independently of others, enabling complex experiments and the exploration of interactions between different defenses, which will be discussed in more detail in Section~\ref{sec:use_cases}.

Attack and defense methods inherit from the \texttt{Defender} and \texttt{Attacker} classes. Subsequently, classes for specific types of attacks and defenses: evasion, poisoning, and privacy are derived from these base classes. 

Thus, to add a new attack or defense method to the registry, it is sufficient to inherit from the appropriate attack or defense class and override the methods of the parent class as needed.

\subsubsection{Interpretation module}

The interpretation module consists of an interpretation manager, which is responsible for saving and loading interpretation results, executing interpretation methods, and calculating interpretation metrics (e.g., fidelity).

The interpretation methods themselves inherit from the \texttt{Explainer} class, which can be overridden to add a new interpretation method to the method registry. If the interpretation method falls under self-interpretable methods and requires specific architectural features of the model, it is necessary to additionally define the constraints of the interpretation method.

After training the model, the user must specify the model manager and the interpretation method to be executed.

\subsection{Frontend}
The web interface of GNN-AID replicates most backend capabilities, allowing users to interact with the tool through a graphical UI in their browser.
The standard workflow includes selecting or building a dataset, constructing or loading a GNN model, and training the model. After that, a user can go three ways.

1. Analyzing graph data. This scenario enables the visualization of the training process. Users can observe detailed intermediate outputs, such as weight matrices from each layer, logits and predictions for each node, and the current performance metrics on training, validation, and test sets.

2. Interpreting GNN models.  In the case of post-hoc interpretation, an interpretation algorithm is specified after model training. It provides either a local or global interpretation. Alternatively, the user can build a self-interpretable model and receive interpretation during the training process.

3. Applying attacks and defenses to GNN models. Similar to the previous scenario, the user specifies the parameters for attacks and/or defenses.

\begin{figure}[h]
    \centering
    \includegraphics[width=0.46\textwidth]{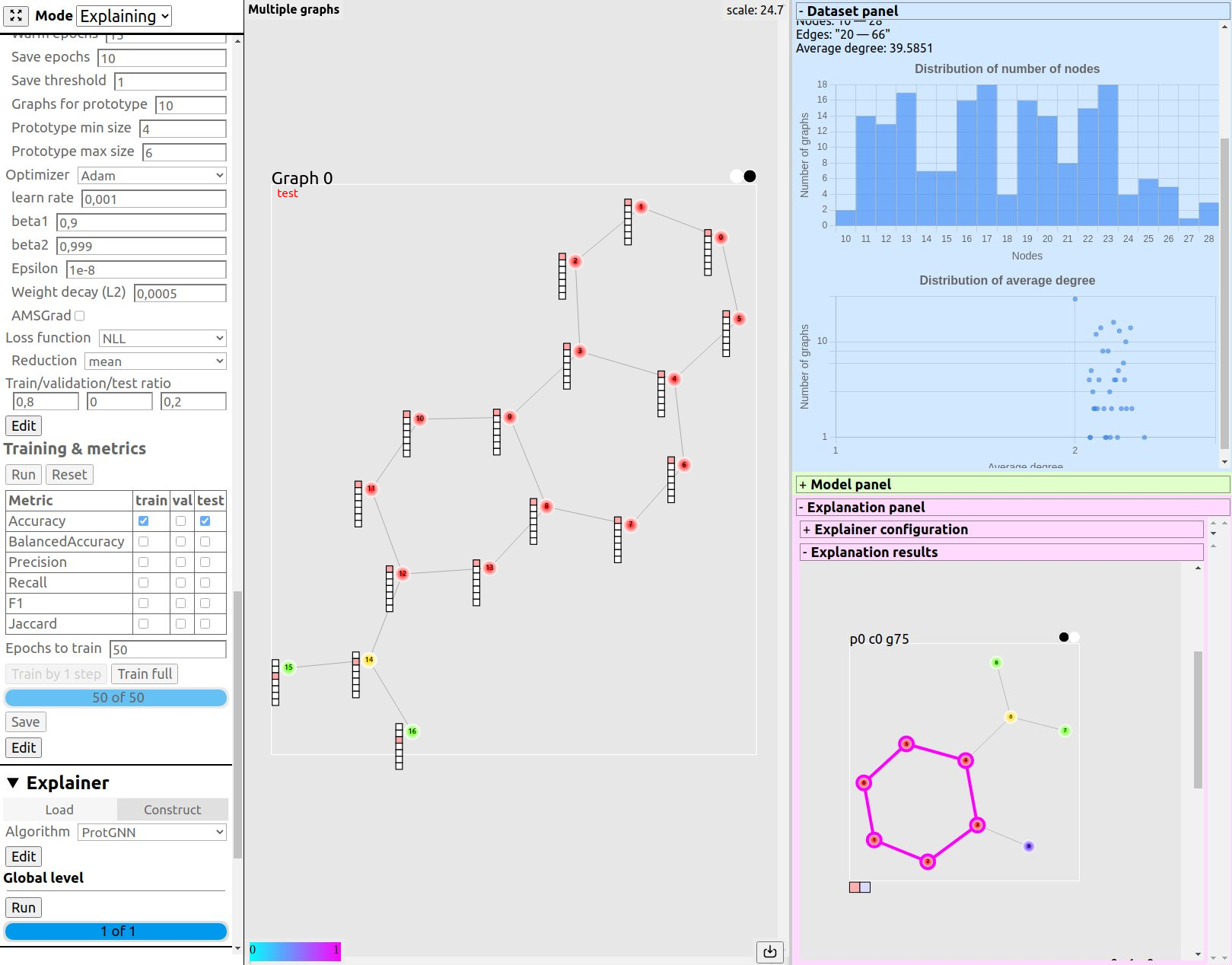}
    \caption{Screenshot of the web interface in interpretation mode. The left side is a menu panel for selecting the dataset (not visible), model training parameter, etc. (additional elements are visible after scrolling down). In the center, there is a visualization of a molecule graph from the MUTAG dataset with additional information around the nodes. The right panel contains some dataset statistics (the upper part), and the global model interpretation via the ProtGNN method. A prototype is shown as a purple highlighted subgraph.}
\label{fig:front-screenshot}
\end{figure}

Figure~\ref{fig:front-screenshot} shows a screenshot for example.

\subsection{Summary}

The modular architecture enables rapid expansion of the framework's supported methods for attacks, defenses, and interpretations. While the interaction between modules is based on PyTorch, all modules rely on core methods that can be overridden by the user. Through an additional interface layer, the framework can be adapted to work with models and methods built on other libraries or even other programming languages, provided the core methods are correctly implemented.

The availability of a web interface allows users with limited expertise in graph-based machine learning methods to interact with the framework effectively. Additionally, the interface provides visualization for each stage of interaction with the framework.

\section{Implemented Methods in the Framework}
\label{sec:methods}
Here we describe interpretation methods, attacks, and defense algorithms implemented in GNN-AID.
\subsection{Interpretation methods}
\subsubsection{GNNExplainer}
GNNExplainer~\cite{ying2019gnnexplainer} identifies a subgraph that maximizes the mutual information between the model's predictions and distribution of possible subgraph structures by training a differentiable mask over edges.

\subsubsection{PGExplainer}
PGExplainer~\cite{luo2020parameterized} uses a parametric model to identify influential edges and nodes by learning probabilities of their contribution to predictions. It employs thresholding on predicted edge probabilities to highlight the most significant graph components.

\subsubsection{PGMExplainer}
PGMExplainer~\cite{vu2020pgm} constructs a probabilistic model to determine important subgraphs influencing specific node predictions. By minimizing the prediction discrepancy between the full graph and subgraphs, it evaluates node and edge importance.

\subsubsection{SubgraphX}
SubgraphX~\cite{yuan2021explainability} uses the Shapley value and Monte Carlo Tree Search (MCTS) to generate a connected subgraph that best explains the model’s predictions. It evaluates subgraph importance based on contribution to model output.

\subsubsection{ZORRO}
ZORRO~\cite{funke2020hard} selects important nodes and features from a computational subgraph by masking components and injecting noise. The algorithm greedily iterates through the nodes and features of the original computation graph and produces a minimal set S that would correspond to the set value of the $fidelity$ metric.

\subsubsection{GraphMask}
The main goal of GraphMask~\cite{schlichtkrull2020interpreting} is to identify edges in a graph that can be removed (considered "redundant") without affecting the model's predictions. The algorithm uses stochastic binary switches $z_{u,v}$ to determine whether to keep the edge $(u, v)$ or replace it with a baseline value on layer $k$: 
$$
\hat{m}_{u,v}^{(k)} = z_{u,v}^{(k)} \cdot m_{u,v}^{(k)} + (1 - z_{u,v}^{(k)}) \cdot b^{(k)}
$$
Each value $z_{u,v}^{(k)}$ is determined by a learned function $g_{\pi} ( h_{u}^{(k-1)}, h_{v}^{(k-1)}, m_{u,v}^{(k)} )$, which is implemented as a neural network. Here $h_{u}^{(k-1)}$, $h_{v}^{(k-1)}$ are the hidden states of the nodes, and $m_{u,v}^{(k)}$ is the original message. $g_{\pi}$ is trained on a training data set, which allows it to identify common patterns in graphs and work not only on the training set but also on new (test) data. 

\subsubsection{ProtGNN}
ProtGNN~\cite{zhang2022protgnn} uses the paradigm of prototype learning to build self-interpretable GNN. This method implies the attachment of the prototype layer to any GNN backbone that stores embeddings of prototype graphs for each predicted class. Firstly, the backbone produces embedding for the input graph, then it is compared with each prototype, and similarity metrics are passed to the linear classifier for producing the output label. As the prototype layer stores just a trainable tensor, for interoperability, each embedding within it is projected into one of the graphs from the train set. This projection is performed by detecting a subgraph that has the most similar embedding to the prototype embedding. Monte-Carlo Tree Search (MCTS) is used for subgraph definition.

\subsubsection{NeuronAnalysis}
NeuronAnalysis~\cite{Han2022GlobalCI} is a post-hoc interpretation method for graph neural networks that focuses on the explanation of distinct neuron behavior within GNN. An algorithm adopts a concept-learning technique of interpretation when a human understanding of logical concepts serves as an explanation. Each concept is a combination of basic concepts with logic operations: NOT, AND, OR. Basic ones can be domain-based (for molecular datasets like MUTAG~\cite{debnath1991structure} it can be ''this node is a carbon atom'') or general concepts for every graph that specifies graph structure (for example, ''this node has degree $>$ 5'').

\subsection{Attack methods}
\subsubsection{RandomPoison}
The RandomPoison is a poisoning attack that randomly removes a specified percentage of edges from a graph to degrade the model’s performance.

\subsubsection{Nettack}
Nettack~\cite{zugner2018adversarial} modifies graph structure and features through a bi-level optimization process: the first level maximizes the impact on the classification of the target node, while the second level minimizes the changes to the graph to ensure they remain as inconspicuous as possible. The attack involves two types of modifications: alterations to the graph structure (such as adding or removing edges) and modifications to the node features (adjusting feature values).

\subsubsection{Fast Gradient Sign Method (FGSM)}
FGSM~\cite{goodfellow2014explaining} is a white-box evasion attack method that follows the idea of changing input in the opposite direction of the gradient of loss function: $$x^{'} = x + \epsilon \  {sign}(\nabla l(f(x),y))$$

\subsubsection{Projected Gradient Descent (PGD)}
PGD~\cite{mkadry2017towards} is an iterative modification of the FGSM white-box evasion attack method. The main difference is that the data shift
is done in several steps. After each step, the gradient signs are recalculated:
$$
x_{i+1}^{'} = \text{Clip}_{\epsilon} \left\{ x_{i}^{'} + \alpha \cdot \text{sign} \left[ \nabla l(f(x),y) \right] \right\}
$$
where $x_{i+1}^{'}$ denotes the changed input data since the previous iteration, $\text{Clip}$\{\} limits the resulting data shift to no more
than $\epsilon$ and $\alpha$ denotes the shift step size at each iteration.\\
In addition to the attack on graph features, the GNN-AID PGD supports an attack on the graph structure. In this case, the goal of the PGD~\cite{xu2019topology} attack is to change the graph structure to maximize the classification error of the target node. After solving the optimization problem using gradient descent, a random sampling method is used to generate discrete perturbations, since changes in the graph must be discrete (e.g. adding or removing edges).

\subsubsection{Q-Attack}
Q-Attack~\cite{chen2019ga} is designed especially for community detection tasks and removes edges that connect nodes with its communities and adds edges to other ones. This pair of operations (remove/add one edge) is called rewiring and is considered a preferable way of attacking edges because it does not change graph statistics such as several edges or degree distribution. 
\par As the method was created for community detection tasks, the fitness function of the genetic algorithm uses modularity of graph, so it requires community labeling. However, in other tasks, it is also possible to use labels that being predicted by GNN. For generalization purposes, GNN-AID uses these predicted answers as labels for modularity calculation and fitness function.

\subsubsection{Contrastive Loss
Gradient Attack (CLGA)}
CLGA~\cite{zhang2022unsupervised} is a graph poison attack method that works in an unsupervised way. Within each iteration, the algorithm produces two augmented views of the graph and treats different views of one node as a positive pair and views of different nodes as a negative pair. Contrastive loss is calculated on these two views and edge to addition/deletion is chosen based on the gradient sign and corresponding value of adjacency matrix. 

\subsubsection{MetaAttack}
MetaAttack~\cite{Zgner2019AdversarialAO} is a poison attack method based on meta-learning. It considers graph matrix structure as a hyperparameter and calculates gradients concerning it. Therefore adjacency matrix is modified edge-by-edge on each iteration of the algorithm. MetaAttack works in the black-box scenario and trains a surrogate model on each iteration for obtaining unknown labels. Computation of meta-gradients may be expensive and authors propose a variant of method with approximate calculation of them. GNN-AID supports both options (MetaAttackFull and MetaAttackApprox respectively).

\subsection{Defense methods}
\subsubsection{Gradient Regularization}
Modifying of loss function used by a model with additional gradient regularization~\cite{finlay2021scaleable} can serve as a defense method too.
$$
L = l(f(x),y) + \lambda (\frac{1}{h^2n}\norm{f(z) - f(x)}_2^2),
$$
where
$$
z = x + h \ \frac{\nabla l(f(x),y)}{\norm{\nabla l(f(x),y)}_2},
$$
$h$ is a quantization step and $\lambda$ is a regularization coefficient.

\subsubsection{Defensive Distillation}
Distillation as a defense method is about creating a copy of the original model that is more robust to attacks. The new model uses the original one as a teacher and so-called smooth labels for this model are obtained using the $softmax(x, T)$ on the last layer of the teacher:
$$
{{softmax}(x,T)}_i = \frac{e^{x_i/T}}{\sum_j {e^{x_j/T}}}
$$

\subsubsection{Adversarial Training}
Adversarial training~\cite{goodfellow2014explaining} is a defense technique that implies adding adversarial examples in a train set. Therefore adversarial part is added to the loss function:
$$
L = l(f(x),y) + \lambda \ l(f(x^{'},y)),
$$
where $x^{'}$ is adversarial sample and $\lambda$ is adversarial training coefficient.

\subsubsection{Data Quantization Defense}
Quantization is a preprocessing technique that transforms continuous values into a discrete set of values arranged on a uniform grid~\cite{guo2017countering}. While this method may diminish the quality of the original data, it can effectively mitigate the effects of adversarial attacks. Consequently, the fault diagnosis model needs to be retrained using the quantized data.

\subsubsection{Autoencoder Defense}
Autoencoders can be used to perform robust training too as it was shown in~\cite{meng2017magnet}. In this case, minimized loss function:
$$
L = \norm{x_{AE} - x}_1,
$$
where $x_{AE} = {autoencoder}(x+\epsilon)$ is a reconstructed data and $\epsilon$ is added noise.

\subsubsection{JaccardDefense}
JaccardDefense~\cite{wu2019adversarial} is a poison defender that removes edges between nodes that are not similar concerning the Jaccard Index (binary features of nodes being compared). This is followed by the idea that many attack methods are trying to connect not similar nodes to shadow important links.

\subsubsection{GNNGuard}
GNNGuard~\cite{zhang2020gnnguard} modifies message-passing functions to diminish the influence of suspicious edges. This is achieved with additional defense weights that are multiplied with passed messages and with node embedding from previous layers. These defense weights are jointly learned with network parameters.

\section{Use Cases}
\label{sec:use_cases}

\subsection{Educational use case}

The GNN-AID framework can be used as an educational tool for beginning GNN learners. It provides real-time visualization of the training process on small demonstration graphs. Users can observe the model’s trainable weights; node features, embeddings, predictions, and true labels directly on the graph. Additionally, the system generates plots of various training metrics, offering deeper insights into the learning dynamics. Screenshots can be found in the appendix. 

\subsection{Developer use case}

The GNN-AID framework can be used to design, test, and experiment with GNN models. It offers modular APIs for custom architectures and datasets, and built-in tools for interpretability, robustness evaluation, and attack/defense analysis. The framework also provides visualization of node embeddings, predictions, and graph-level statistics, streamlining workflows.

\subsection{Researcher use case}

The GNN-AID framework can be used to study the interactions between various attack and defense types in GNNs. It enables researchers to explore the relationship between model interpretability and robustness. The framework includes MLOps tools for ensuring reproducibility and versioning of results. Additionally, it can be used to create benchmarks for comparing interpretation methods, attack strategies, and defense mechanisms in graph-based models.

\section{Conflicting interactions among protection mechanisms}

As demonstrated in \cite{szyller2023conflicting}, defense methods can conflict in the image domain. In this work, we further investigate whether such conflicts are characteristic of defense methods in the graph domain. Unlike images, attacks and defenses on graphs can modify not only features but also the graph structure.

For the experiment, the following defense methods were selected: JaccardDefense (Jaccard), Adversarial Training (AdvTrain), and Gradient Regularization (GradGeg). The first method is designed to defend against poisoning attacks, while the other two target evasion attacks. The attacks considered were PGD as an evasion attack and CLGA as a poisoning attack. The experiments were conducted on the Cora dataset. Detailed information on the hyperparameters of the attack and defense methods is provided in the supplementary materials.

According to the training pipeline described in Algorithm \ref{alg:training_pipeline}, with the specified attack and defense methods, the poisoning attack was applied first, followed by the poisoning defense method. During training, a defense against evasion attacks was added, and before model evaluation, an evasion attack was applied to the test data.

The experiment examined various combinations of attack and defense methods of both types. Cases where a defense against one type of attack was added while the only attack was of the other type were not considered.

The results of the experiments were averaged over 15 runs.

\begin{table}[h!]
\centering
\begin{tabular}{cccc}
\toprule
\textemdash & No EvDefense & AdvTrain & GradReg \\
\midrule
No PoisDefense & $90 \pm 0$ & $88 \pm 2$ & $91 \pm 1$ \\
Jaccard   & $82 \pm 5$ & $81 \pm 5$ & $80 \pm 5$ \\
\bottomrule
\end{tabular}
\caption{Accuracy of a model with different combinations of defense methods when not under attack on Cora dataset}
\label{tab:gcn_cora_no_attack}
\end{table}

\begin{table}[h!]
\centering
\begin{tabular}{cccc}
\toprule
\textemdash & No EvDefense & AdvTrain & GradReg \\
\midrule
No PoisDefense & $57 \pm 3$ & $78 \pm 3$ & $70 \pm 4$ \\
Jaccard   & $46 \pm 4$ & $69 \pm 5$ & $63 \pm 2$ \\
\bottomrule
\end{tabular}
\caption{Accuracy of a model with different combinations of defense methods when model attack only by evasion atttack PGD on Cora dataset}
\label{tab:gcn_cora_pgd}
\end{table}

\begin{table}[h!]
\centering
\begin{tabular}{cccc}
\toprule
\textemdash & No EvDefense & AdvTrain & GradReg \\
\midrule
No PoisDefense & $67 \pm 2$ & \textemdash & \textemdash \\
Jaccard & $76 \pm 6$ & $67 \pm 3$ & $61 \pm 2$ \\
\bottomrule
\end{tabular}
\caption{Accuracy of a model with different combinations of defense methods when model attack only by poison attack CLGA on Cora dataset}
\label{tab:gcn_cora_clga_no_attack}
\end{table}

\begin{table}[h!]
\centering

\begin{tabular}{cccc}
\toprule
\textemdash & No EvDefense & AdvTrain & GradReg \\
\midrule
No PoisDefense & \textemdash & \textemdash & \textemdash \\
Jaccard & \textemdash & $57 \pm 4$ & $42 \pm 1$ \\
\bottomrule
\end{tabular}
\caption{Accuracy of a model with different combinations of defense methods when model attack by PGD and CLGA together on Cora dataset}
\label{tab:gcn_cora_clga_pgd}
\end{table}

From Tables \ref{tab:gcn_cora_pgd} and \ref{tab:gcn_cora_clga_no_attack}, it is evident that defense methods against evasion attacks significantly improved the model's performance under PGD attacks. Similarly, the defense method against poisoning attacks effectively mitigated the negative impact of the CLGA attack.

However, as shown in Tables \ref{tab:gcn_cora_no_attack}, \ref{tab:gcn_cora_pgd}, \ref{tab:gcn_cora_clga_no_attack}, and \ref{tab:gcn_cora_clga_pgd}, combining defense methods degraded the overall accuracy metric in all cases. Furthermore, in each pair, adding a defense method unrelated to the type of attack being conducted worsened the model's performance. In scenarios involving combined attacks with both methods, none of the defense combinations effectively countered the combined threat.

This experiment summarizes that the graph domain is also susceptible to conflicts between defense mechanisms, similar to the image domain. Additionally, this highlights the importance of studying not only individual trustworthiness criteria but also the ability to ensure their compatibility. Without such studies, the deployment of AI systems in critical infrastructure will remain highly uncertain.

\section{Conclusion and Future Work}
\label{sec:conclusion}

We created a framework, GNN-AID, that provides methods for analyzing, interpreting, and defending Graph Neural Networks (GNNs) while addressing key aspects of Trusted AI: interpretability and robustness. By combining ease of use, access to no-code solutions, API, and visualization tools with MLOps support for reproducibility, GNN-AID enables new opportunities for researchers and developers alike.

In this article, we used GNN-AID to demonstrate for the first time the intricate relationship between defense strategies against evasion and poisoning attacks in the context of graph data. This finding underscores the complexity of designing effective defenses and highlights the need for a holistic approach to building trusted GNN and other AI models.

GNN-AID is a versatile tool that fosters scientific exploration and industrial solutions for graph data analysis, advancing the principles of Trusted AI.

The main direction for future work is to expand the range of available attacks, defenses, and interpretation methods. Additionally, this includes introducing a class of privacy attacks and corresponding defense mechanisms.


\bibliographystyle{named}
\bibliography{reference}

\end{document}


\maketitle


%


\section{Frontend examples}

\begin{figure*}[ht]
    \centering
    \includegraphics[width=\textwidth]{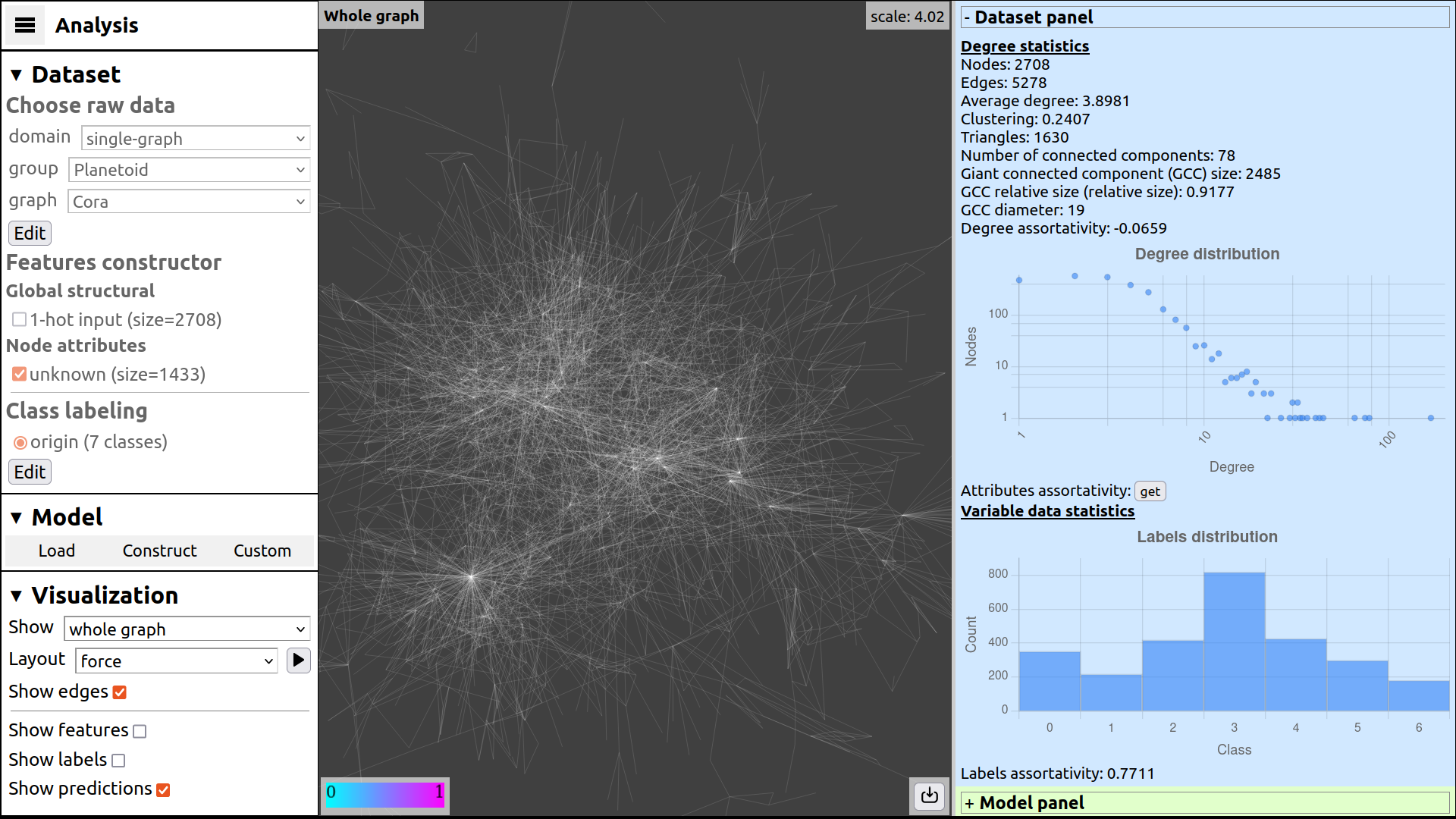}
    \caption{
    Screenshot of dataset statistics for Cora citation graph. Graph is zoomed out so its nodes are not shown.
    }
\label{fig:front-cora}
\end{figure*}

\begin{figure}[ht]
    \centering
    \includegraphics[width=0.5\textwidth]{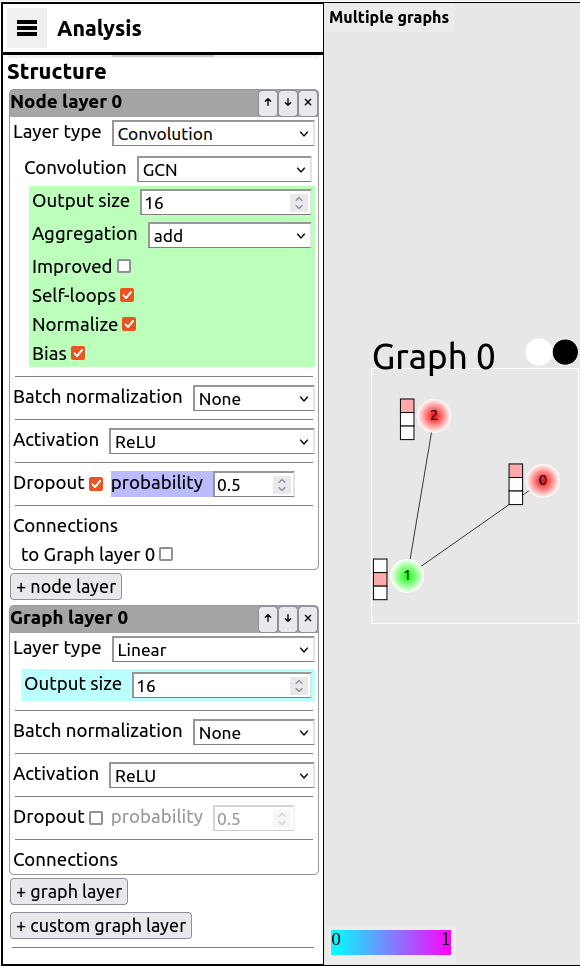}
    \caption{
    Screenshot of model constructor menu. The model structure in this example starts with a Node layer with a layer of type Convolution, named GCN and output size 16, followed by a batch normalization, ReLU activation unit, and dropout with 0.5 probability. User can add, move, or remove node layers by pressing buttons. The second layer here is a Graph layer of linear type. Below is a button '+ custom graph layer' that adds (currently supported only 1 option) a prototype layer which makes model self-interpretable.
    }
\label{fig:front-model-constr}
\end{figure}

\begin{figure*}[ht]
    \centering
    \includegraphics[width=\textwidth]{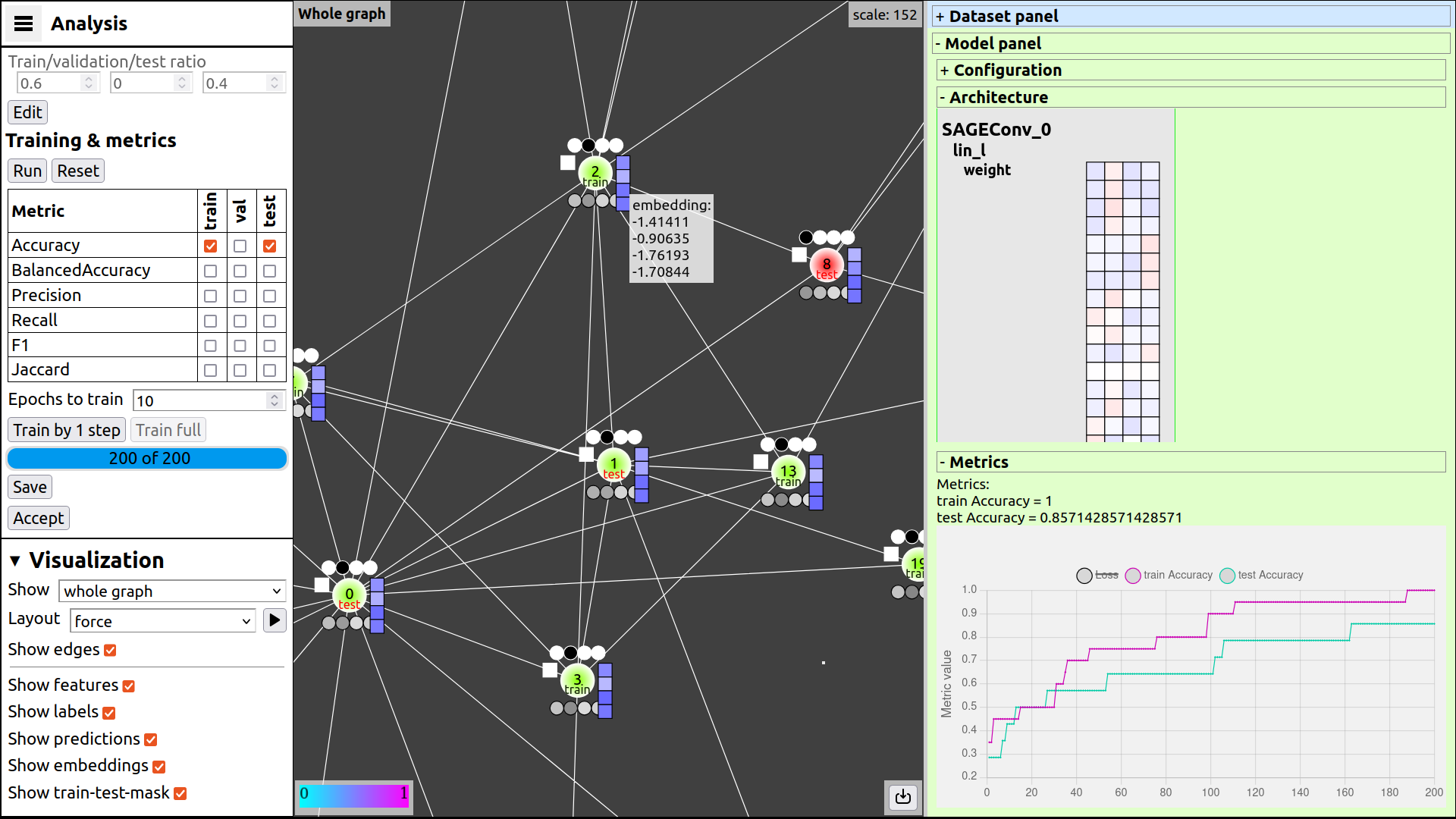}
    \caption{
    Screenshot of the web interface in analysis mode, suit for introducing beginners to the functionality of GNN models using the karate-club graph as an example. On the left, a menu panel displays model training parameters and visualization control flags for the graph shown in the center. The central section contains graph nodes and edges additional node information: true labels (one of four classes) at the top, normalized prediction vectors (0-1) at the bottom, representation vectors on the right, and a square on the left that reveals feature vectors upon hover (hidden here due to size). Node color indicates its class. On the right, the Model panel displays the model architecture with weights for each layer (numerical values appear on hover over the squares) and a metric plot below, showing their evolution over 200 training epochs.
    }
\label{fig:front-edu}
\end{figure*}

\begin{figure*}[ht]
    \centering
    \includegraphics[width=\textwidth]{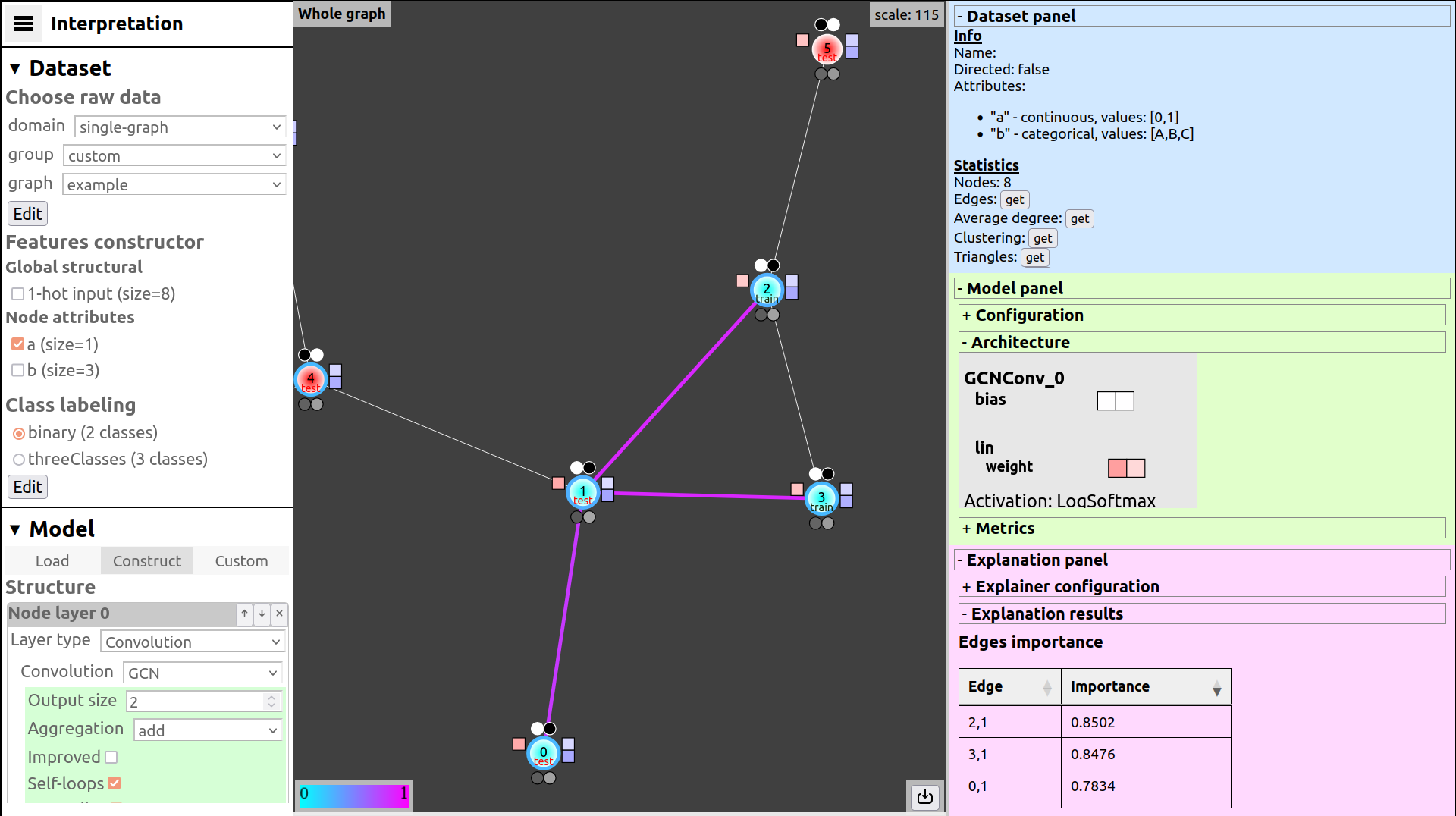}
    \caption{
    Screenshot of the web interface in interpretation mode. On the left, a menu panel allows selecting the dataset, model (partially visible), and other options (additional elements appear when scrolling down). In the center, a graph visualization is displayed with additional information around the nodes and the results of explaining the classification for node 1, shown as purple edges in its neighborhood. On the right, information panels are arranged from top to bottom: about the dataset, the model, and the interpretation.
    }
\label{fig:front-interpret1}
\end{figure*}

\begin{figure*}[ht]
    \centering
    \includegraphics[width=\textwidth]{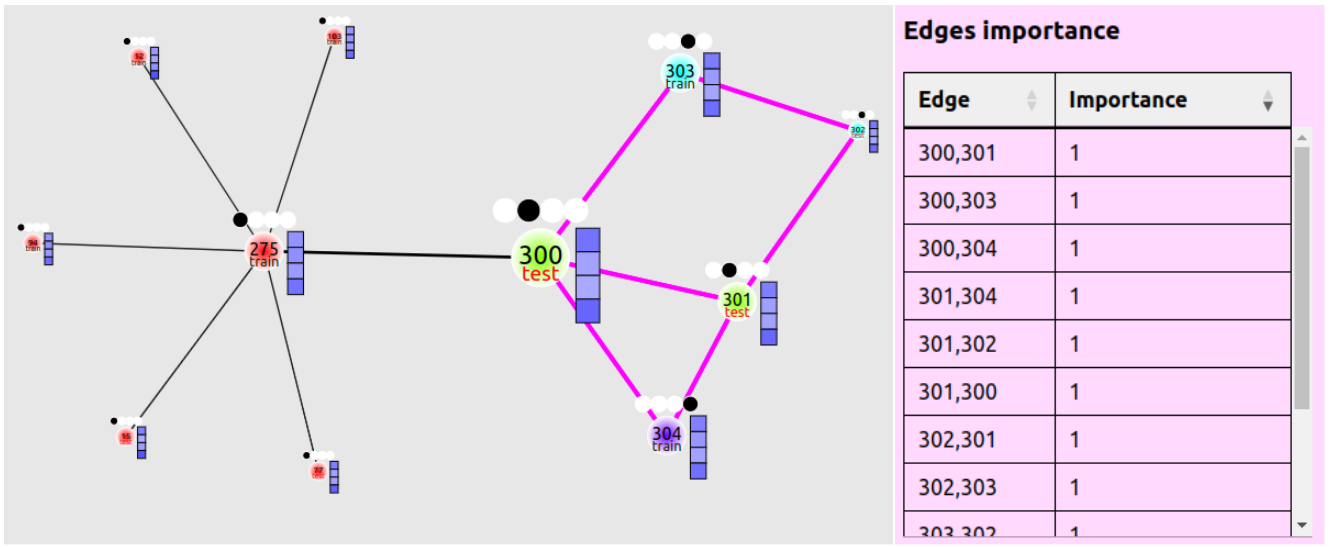}
    \caption{
    Result of the PGExplainer algorithm on the BAShapes dataset\protect\footnotemark. Interpreting the model's prediction for node 300. The algorithm correctly identifies influential edges in the neighborhood.
    }
\label{fig:front-interpret2}
\end{figure*}

Here we provide several screenshots of UI demostrating frontend work. Figure~\ref{fig:front-cora} shows dataset statistics exploration on Cora citation graph\footnote{\url{https://paperswithcode.com/dataset/cora}}.
Figure~\ref{fig:front-model-constr} shows no-code model constructor.
Figure~\ref{fig:front-edu} shows educational example.
Figures~\ref{fig:front-interpret1} and~\ref{fig:front-interpret2} show interpretation examples.

\footnotetext{\url{https://pytorch-geometric.readthedocs.io/en/latest/generated/torch_geometric.datasets.BAShapes.html}}

\section{Conflicting interactions among protection mechanisms; 
Experimental parameters}

Experiment parameters, including training parameters, model architecture, and hyperparameters of attack and defense algorithms.

The two-layer GCN model (GCN-2l) and the two-layer GIN model (GIN-2l) were trained as models. 
\begin{verbatim}
GCN-2l(
  Sequential(
    (0): GCNConv(input_size, 16)
    (1): ReLU(inplace)
    (2): GCNConv(16, output_size)
    (3): LogSoftmax(inplace)
  )
)
\end{verbatim}

\begin{table}
    \centering
    \begin{tabular}{@{}ll@{}}
        \toprule
        \textbf{Method} & \textbf{Hyperparameters} \\ \midrule
        CLGA & 
        \begin{tabular}[c]{@{}l@{}}
            learning\_rate = $0.01$ \\
            num\_hidden = $256$ \\
            num\_proj\_hidden = $32$ \\
            activation = $prelu$ \\
            drop\_edge\_rate\_1 = $0.3$ \\
            drop\_edge\_rate\_2 = $0.4$ \\
            tau = $0.4$ \\
            num\_epochs = $3000$ \\
            weight\_decay = $1e-5$ \\
            drop\_scheme = $degree$ 
        \end{tabular} \\ \midrule
        
        PDG & 
        \begin{tabular}[c]{@{}l@{}}
            $\epsilon = 0.005$ \\
            $num\_iterations = 10$ \\
            $\alpha (learning_rate) = 0.0005$
        \end{tabular} \\ \midrule
        
        Jaccard & 
        \begin{tabular}[c]{@{}l@{}}
            threshold = $0.4$ 
        \end{tabular} \\ \midrule
        
        Adversarial training & 
        \begin{tabular}[c]{@{}l@{}}
            attack\_name = $FGSM$ \\
            $\epsilon = 0.01$ 
        \end{tabular} \\ \midrule
        
        Gradient Regularization & 
        \begin{tabular}[c]{@{}l@{}}
            regularization\_strength = $50$ 
        \end{tabular} \\ 
        
        \bottomrule
    \end{tabular}
    \caption{Hyperparameters of attack and defense methods}
    \label{tab:hyperparams}
\end{table}

Cora dataset parameters: 2708 nodes, 10556 edges, 7 classes, 1433 features.

Hyperparameters of attack and defense methods are presented in Table \ref{tab:hyperparams} 

Training parameters: Adam optimizer with default parameters and NLLLoss error function from PyTorch library were used for training, no batch splitting was used. The number of training epochs was 200 to achieve high classification accuracy on all datasets. The data was split in a ratio of 80 to 20 for training and testing, respectively.

The main metric for evaluating the models was the Accuracy metric on the test data after applying all declared attack and defense methods or on the original data if no attack or defense was declared.